\pdfoutput=1

\documentclass[11pt]{article}

\usepackage[preprint]{acl}

\usepackage{times}
\usepackage{latexsym}

\usepackage[T1]{fontenc}

\usepackage[utf8]{inputenc}

\usepackage{microtype}

\usepackage{inconsolata}

\usepackage{inconsolata}
\usepackage{graphicx,multirow}

\title{ChartCheck: Explainable Fact-Checking over Real-World Chart Images}

\author{Mubashara Akhtar,\textsuperscript{1} Nikesh Subedi,
\textsuperscript{2} Vivek Gupta,\textsuperscript{3} Sahar Tahmasebi,\textsuperscript{4} \\ {\bf Oana Cocarascu\textsuperscript{1}} and {\bf Elena Simperl\textsuperscript{1}} \\[5pt]
        \textsuperscript{1}King's College London
        \textsuperscript{2}University of Utah
        \textsuperscript{3}University of Pennsylvania\\[5pt]
        \textsuperscript{4}TIB – Leibniz Information Centre for Science and Technology\\[5pt]
        \texttt{mubashara.akhtar@kcl.ac.uk}} 

\begin{document}
\maketitle
\begin{abstract}

Whilst fact verification has attracted substantial  interest in the natural language processing community, verifying misinforming statements against data visualizations such as charts has so far been overlooked. Charts are commonly used in the real-world to summarize and communicate key information, but they can also be easily misused to spread misinformation and promote certain agendas. In this paper, we introduce ChartCheck, a novel, large-scale dataset for explainable fact-checking against real-world charts, consisting of 1.7k charts and 10.5k human-written claims and explanations. We systematically  evaluate ChartCheck using vision-language and chart-to-table models, and propose a baseline to the community. Finally, we study chart reasoning types and visual attributes that pose a challenge to these models.\footnote{Code and data available at \url{https://github.com/mubasharaak/ChartCheck}}

\end{abstract}

\section{Introduction}

\begin{figure}
\centering
\includegraphics[scale=0.35]{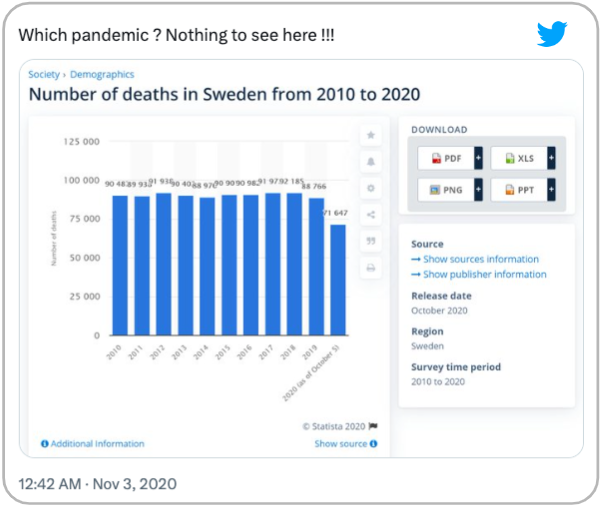}
\caption{\label{fig:chart_misinfo_sample} Twitter post related to COVID misinformation~\citep{LisnicPLK23}.}
\vspace{-1.5em}
\end{figure}

Data visualizations (e.g. bar charts, pie charts, line graphs) are common in real-world data sources and are frequently used in scientific documents, textbooks, news articles, and social media to summarize key information in a visual form, 
information which is often not fully repeated in the associated text~\citep{CarberryED2006}. 
However, data visualizations are also commonly misused to spread misinformation,\footnote{See Fig.~\ref{fig:chart_misinfo_sample} and appendix for examples.} 
being included in political and business advertisements, or spread on social media, with the aim to convince the audience towards certain agendas~\citep{LoGSWBQ22, LisnicPLK23}. 
For example, during the COVID pandemic, the term \textit{``counter-visualizations''} was coined for charts 
circulating 
on social media platforms to counter COVID-related health measures~\citep{LeeYIJS21} (see Fig.~\ref{fig:chart_misinfo_sample}). 
Charts were also (mis-)used by Brexit campaigners during the referendum.\footnote{See Fig.~\ref{fig:chart_misinfo_sample2} and ~\ref{fig:chart_misinfo_sample3} in the appendix.}

\begin{figure}
\advance\leftskip-0.2cm
    \begin{tabular}{ c c }
        \fbox{\begin{minipage}{19.1em}
                \small       
                    \centering
                \textbf{\underline{Evidence}}\\
               \raggedright{\textbf{Chart:}} \\
                    \centering
                    \includegraphics[scale=0.26]{figures/chartcheck_example.png}\\
               \raggedright{\textbf{Caption:} $2006$ Mexican Presidential election vote count progression. Percentage of poll stations counted vs. percentage of candidate votes.} \\
               \rule{\linewidth}{0.01em}
               \textbf{Claim:} The percentage of votes for Loprez Obrador decreased over time as more poll stations were counted.\\
                \textbf{Explanation:}
                The chart shows a generally downward trend in the percentage of votes for Obrador as more poll stations were counted. As more votes were counted, the other candidate gained a larger percentage of the overall vote. \\
                
        \end{minipage}} & 
    \end{tabular}
    \caption{A dataset sample labelled as \textit{supported}.}
    \label{fig:chart_sample}
    \vspace{-1.5em}
\end{figure}

While previous research has focused on misleading chart design (i.e. truncated axis)~\citep{LoCYQ24, HemsleySnyder18}, a less studied but prevalent issue is misleading statements that exploit information fallacies during chart interpretation. 
Verifying these statements requires extracting information from charts that use tightly integrated text and visual elements to represent information.
These visual elements consist of various lines and shapes, and have different colors, scales, angles, and orientations. 
Moreover, various reasoning skills, including numeracy and language understanding, are required to successfully verify potentially misleading statements against charts. 



In this paper, 
we introduce \emph{ChartCheck}, a novel dataset for explainable fact-checking against data visualizations. 
ChartCheck is the first large-scale dataset with $1.7k$ real-world charts extracted from the web and $10.5k$ human-written claims and explanations (see Fig.~\ref{fig:chart_sample} for an example).
Different to previous datasets, ChartCheck provides explanations as justifications for claim verification. 
Explaining fact-checking decisions is an important aspect 
as a major goal of fact-checking is to convince readers, and debunking misinformation by only labelling it as false is often not sufficient~\citep{guo-etal-2022-survey}.
To collect the data, we apply a four-step crowdsourcing pipeline to filter out noisy images, write claims and explanations, evaluate the data, and apply a final verification check through expert annotators.

We evaluate state-of-the-art (SOTA) models in a finetuning, few- and zero-shot setting. 
Hereby, we consider two baseline architectures: 
$(1)$ vision-language models (VLMs) and 
$(2)$ chart-to-table architectures with separate models for each sub-problem: chart translation (chart-to-table), claim verification, and explanation generation.
Our best-performing chart-to-table baseline reaches $73.8$ accuracy and lags far behind human performance.


We annotate a subset of ChartCheck with $(i)$ reasoning types humans apply for chart understanding~\citep{AmarES05} and $(ii)$ visual attributes common in charts.
We evaluate which reasoning types and chart attributes pose a challenge for SOTA models,
as well as common failure types related to explanation generation. We evaluate model-generate explanations for factuality, semantic coverage, coherence, readability, and redundancy.
Overall, our results show that models generate very fluent, readable and coherent text, but show major limitations w.r.t. understanding and reasoning over charts to generate correct explanations. 
Hence, ChartCheck is 
a challenging problem and will stimulate progress on fact-checking against charts.

To summarise, our \textbf{contributions} are as follows: 

\noindent \emph{1)}    We propose ChartCheck, the first chart dataset for explainable fact-verification with $1.7k$ real-world charts and $10.5k$ human-written claims and explanations. 

\noindent \emph{2)} We evaluate chart-to-table models and VLMs in a finetuning and few-/zero-shot setting. 

\noindent \emph{3)} We study chart reasoning types and visual attributes that pose a challenge to SOTA models, as well as failure types related to explanations. 

\section{The Need for Chart Fact-checking}

\noindent\textbf{Charts in the real-world.}
Data visualizations are commonly consumed on a daily basis, 
e.g. in news articles, text books, scientific papers, and on the Internet~\citep{LoGSWBQ22}. 
Charts, diagrams, plots and infographics are useful tools to summarize and communicate key information in a visual form, 
which is often not fully repeated in the associated text~\citep{CarberryED2006}. 
Moreover, data visualizations are popular data sources for studying data-centric problems in  different domains such as finance, science, health, climate-change. 
For example, during the  pandemic, charts were widely used to guide policymakers in deciding health policies and  communicate COVID information to  the general public (Johns Hopkins University's  coronavirus dashboard is a popular example).\footnote{\url{https://coronavirus.jhu.edu/map.html}}


\noindent\textbf{Misinformed by charts.}
Despite their useful applications, data visualizations are also commonly misused to spread misinformation. 
Previous research on chart-related misinformation focused on misinformation caused by misleading chart design, in particular on visual manipulation techniques, such as  truncated or double axes, missing legends, linear scale for exponential data, confusing colors, or violation of guidelines and best practises related to visualization design~\citep{LoGSWBQ22}.
%
However, a misinformation type related to data visualizations that is less studied but prevalent in the real-world, are information fallacies during chart interpretation.
Analyzing Twitter posts that contain data visualizations, \citet{LisnicPLK23} found that the most common way people mislead with charts is through misleading arguments that exploit information fallacies as opposed to misleading chart design.
They identify common components of misleading chart interpretation such as cherry-picking of the data points and time frames, setting of arbitrary data thresholds, and causal errors, among others.

\noindent\textbf{Reasoning over charts.}
Misinformation related to chart interpretations cannot be identified through visual manipulation detectors but requires multiple sub-steps: $(i)$ understanding the chart content given its context (e.g. caption, surrounding information); $(ii)$ understanding the accompanying claim; $(iii)$ identifying and performing necessary reasoning steps (e.g. arithmetic); $(iv)$ deciding and explaining why the claim is correct or not. 
 %
There are certain challenges unique to reasoning and verification over data visualizations. 
To verify a claim, information needs to be extract from chart images 
that convey meaning through various types of
visual elements such as lines, shapes, different colors, scales, angles, orientations, etc. \citep{LeeJTHLEKWCT2022, liu2023}. 
Moreover, various skills, including numeracy, causal reasoning, and understanding of semantics, need to be successfully applied to verify the given claim. 
While there are many prior works on visual-language misinformation, these have mostly focused on datasets with natural images where the language and visual information is not strongly integrated, e.g. manipulation or out-of-context detection~\citep{akhtar-etal-2023-multimodal}.

\section{ChartCheck: Dataset Creation}

\begin{figure}
\centering
\includegraphics[scale=0.22]{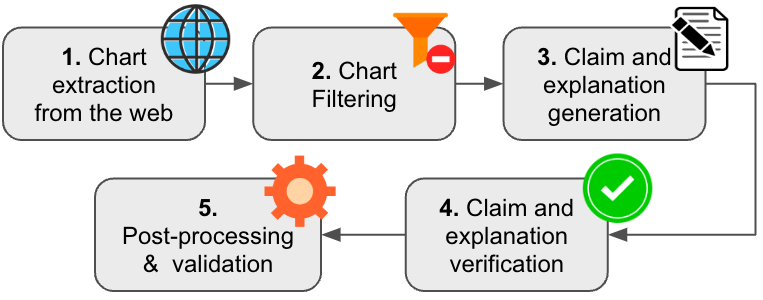}
\caption{\label{fig:datasetpipeline} Overview of the ChartCheck dataset pipeline.}
\vspace{-1.5em}
\end{figure}

Fig.~\ref{fig:datasetpipeline} provides an overview of the dataset creation pipeline. Starting with chart extraction from the web (step $1$), we applied three crowdsourcing tasks (step $2-4$), before post-processing and validating the collected data (step $5$).

\subsection{Data Collection (Step $1$)}

To collect a diverse set of chart images, we used Wikimedia. Commons\footnote{\url{https://commons.wikimedia.org/wiki/Commons:Welcome}}
We extracted horizontal/vertical bar charts, line/area charts, pie/donuts charts, and scatterplots, if the chart description was in English, resulting in 
%
$2,498$ charts.\footnote{See Appendix~\ref{ssec:dataset_details} for further details.}  

\subsection{Crowdsourcing (Steps $2- 4$)} 
\label{sec:crowdsourcing}

We used Mechanical Turk for crowdsourcing.\footnote{Details on worker training/recruitment in Appendix~\ref{ssec:crouwdsourcing_appendix}.} 

\paragraph{Chart filtering (Step $2$).}
First, we filter out noisy chart images for the subsequent tasks. We asked annotators to label if charts were $(i)$ non-English, $(ii)$ not readable, $(iii)$ not understandable (e.g. because crucial information was missing).
We implemented automated checks required 
to submit a task, e.g. if all 
labels were set and annotators spent a minimum time span on each chart.
Moreover, we included in each taskset of seven chart images two golden samples pre-labeled by the authors. Annotators could only submit a taskset if they labelled the gold samples correctly.
Finally, we filtered the collected data using majority voting, resulting in $1,684$ charts.

\paragraph{Claim and explanation generation (Step $3$).}
Claims and explanations were written by a different group of annotators. For each chart, we asked annotators to write one claim supported by the chart and one refuted. 
The workers also wrote a text explaining why the claim is supported/refuted and outlining the reasoning steps. 
To improve the submission quality and guarantee non-superficial claims and explanations, we implemented automated checks and a manual verification step. 
We automatically evaluated the submissions and rejected those that violated the following conditions:
$(i)$ claims had a length between $7-30$ words and explanations between $10-100$ words; $(ii)$ no duplicate claims or explanations were submitted; $(iii)$ claims/explanations did not contain uncertainty terms (e.g. \textit{``perhaps''}, \textit{``maybe'}', \textit{``occasionally''}); $(iv)$ refuting claims were not simple negations of supporting claim; $(v)$ minimum time spent on each chart was $5$ seconds. 
Moreover, we instructed workers to refrain from writing simple claims and gave corresponding examples. 
Finally, we randomly sampled and manually evaluated 
the quality of
one claims-explanation pair for each submitted task before accepting the annotation work.
%
In total, we collected $9,300$ claims and explanations.
On average, claims and explanations have a length of $12.5$ and $23.4$ tokens, respectively. 

\paragraph{Validation of claims and explanations (Step $4$).}
Next, we asked another group of workers to verify the claims and explanations. 
If the assigned claim label did not match its content, workers corrected them. Similarly, workers corrected explanations if they did not explain correctly why a claim was supported/refuted by the chart and its caption. For example, workers identified and corrected partial explanations as well as errors in the explanation text.
We created sets of seven tasks out of which two where gold-labelled, and included automated checks similar to Step $3$. 
We calculated inter-annotator agreement (IAA) scores using Randolph’s kappa  
for 
claim filtering, claim verification, and explanation verification tasks (see Table~\ref{tab:agreement}). 
Agreement over $61.0$ indicates substantial agreement while $51.4$ signifies moderate agreement~\citep{landis1977measurement}. 

\subsection{Post-processing and validation (Step $5$)}

Finally, 
a group of annotators consisting of post-graduate computer science students evaluated all $1.9k$ testset claims and explanations (Table~\ref{tab:dataset_stats}). 
We instructed them to evaluate and (if necessary) correct $(i)$ writing errors (e.g. typos, grammatical errors), $(ii)$ claim labels, $(iii)$ mismatch between claims and explanations. They also $(iv)$ adjusted ``simple'' claims, which require only one reasoning type for verification, to make them more complex.

\begin{table}
\centering
\scalebox{0.79}{
\begin{tabular}{l rrrr}
\hline  
\bf Set &\bf \#Images &\bf \#Claims &\bf Support &\bf Refute \\ \hline
\textbf{Train} & 1,532 & 7,607 & 3,871 & 3,736 \\ 
\textbf{Valid} & 672 & 953 & 494 & 459 \\ 
\textbf{Test $t1$} & 669 & 939 & 487 & 452 \\  
\textbf{Test $t2$} & 151 & 981 & 503 & 478 \\ 
\hline
\textbf{Total} & 1,683 & 10,480 & 5,355 & 5,125\\
\hline
\end{tabular}}
\caption{\label{tab:dataset_stats} Overview ChartCheck statistics. Whilst claims are unique in each set, images may overlap 
except for 
$t2$ that contains images not present in any other split.}
\end{table}


\begin{table}
\centering
\scalebox{0.79}{
\begin{tabular}{l c}
\hline  
\textbf{Task} &\bf R-$\kappa$ \\ \hline
Chart filtering & 66.2\\
Claim verification & 61.5\\
Explanation verification & 51.4\\
\hline
\end{tabular}}
\caption{\label{tab:agreement} Randolph's Kappa (R-$\kappa$) as IAA scores.}
\vspace{-5mm}
\end{table}

\section{Analysis of ChartCheck}

\subsection{Dataset Statistics}

An overview of the ChartCheck dataset is given in Table~\ref{tab:dataset_stats}.
On average, the dataset contains six (three supporting and three refuting) claims per chart image. 
ChartCheck includes two testsets: 
while $t1$ is a testset 
has a similar distribution as the training set, for $t2$ we only included charts which are not part of the training set to
evaluate models' performance on charts 
not seen 
during training. 


\subsection{Chart Attributes}


Following previous fact-checking datasets, e.g.~\citet{DBLP:conf/iclr/ChenWCZWLZW20}, we manually annotated a subset of $100$ charts (i.e. $50$ per testset) with the following attributes: 
chart type;
complexity of charts;
presence of datapoint labels, legends, axes titles, and main title;
axes scales;
background grid characteristics;
number of charts per image;
font size of title, labels, and other text occurring in the chart image;
visual characteristics (e.g. colors and positions of legends).
Fig.~\ref{fig:charttypes} shows the chart types found in the annotated subset.
Around half of the charts ($47.5\%$) include labels their data points
and 
$38.4\%$ have a legend describing the data categories. Most of these charts use colors ($32.3\%$) to group categories and only a small portion ($5.1\%$) uses different patterns.
Charts that have 
an orientation (e.g. bar charts and line charts) 
are 
more often horizontal ($63.2\%$) than vertical ($36.8\%$). 
%
$7\%$ of images have multiple charts depicted in one image.

\subsection{Reasoning with Charts}

\begin{figure}
\centering
\includegraphics[scale=0.15]{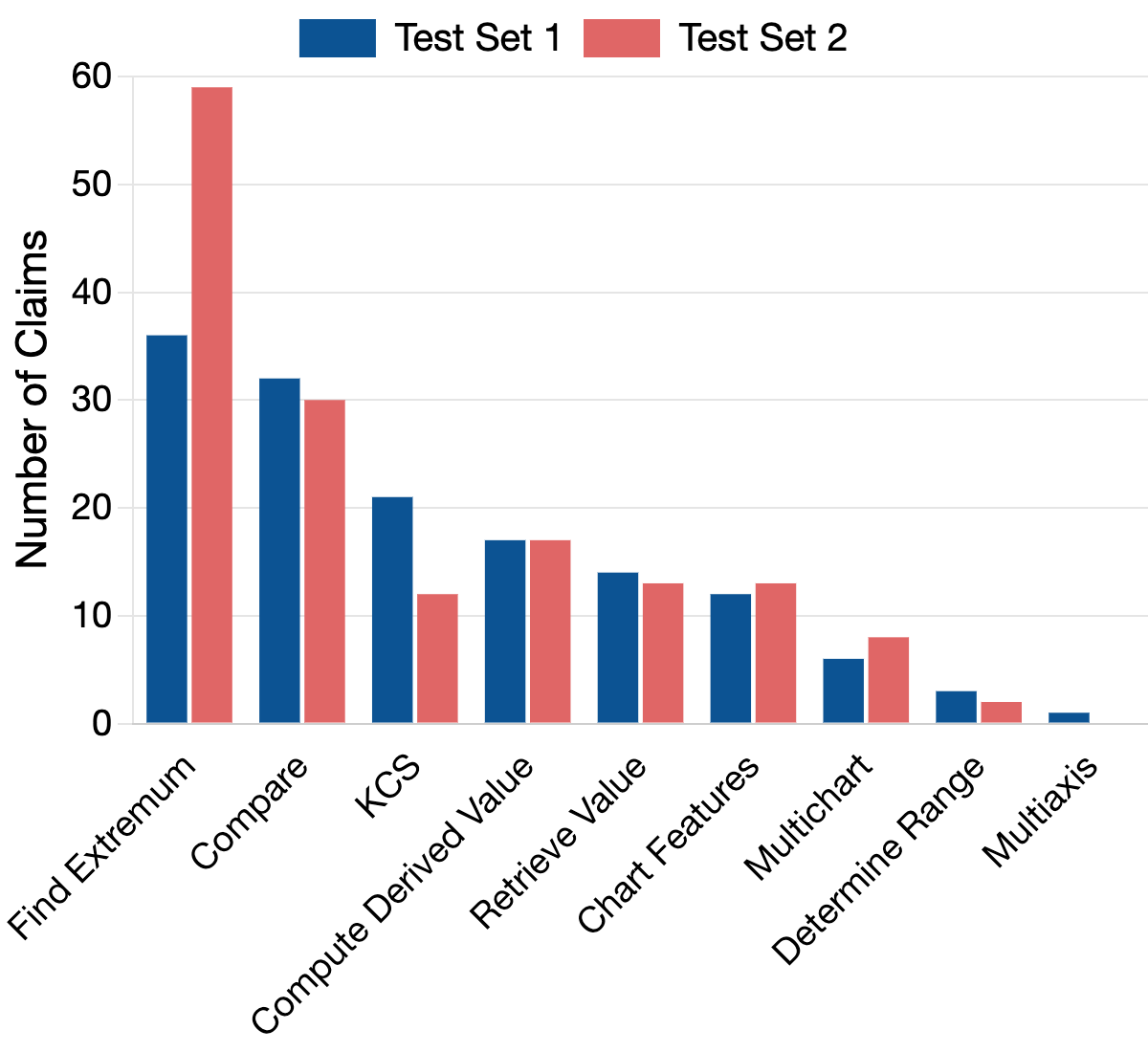}
\caption{\label{fig:reasoning_count} Reasoning types in annotated subset. }
\vspace{-5mm}
\end{figure}

We labelled $230$ dataset samples from both testsets with chart reasoning types based on a taxonomy of reasoning types humans use while interacting with chart data~\citep{AmarES05}. 
Nine different types are present in the labelled subset: find extremum, comparison, world knowledge \& commonsense (KCS), compute derived value, retrieve value, chart features, multichart reasoning, determine range, and multiaxis. Fig.~\ref{fig:reasoning_count} shows the reasoning types per testset. Around $100$ out of all labelled claims require two to four reasoning types for correct classification of claims. 

\subsection{Model-generated Explanations}

We further evaluated explanations by the best-performing, multitask model (i.e. GPT$4$V) in detail. 
We asked postgraduate 
students to annotate $100$ explanations based on the following five dimensions: $(i)$ factuality, $(ii)$ coverage, $(iii)$ readability, $(iv)$ coherence, and $(v)$ redundancy.

\begin{figure*}
\centering
\includegraphics[scale=0.4]{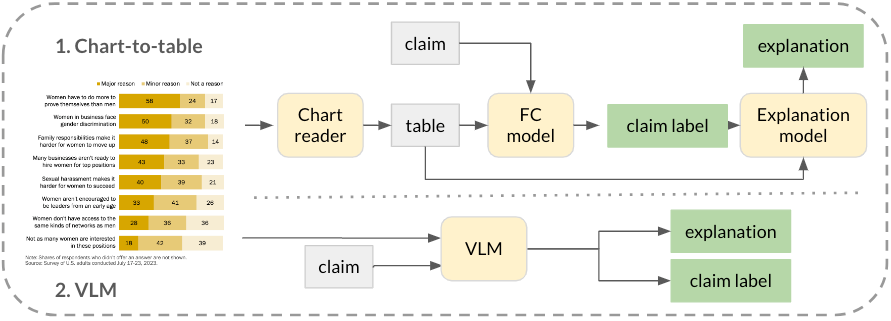}
\caption{\label{fig:baselines} Architecture of chart-to-table and vision-language model (VLM) baselines.}
\end{figure*}

\begin{table*}[h]
\centering
\scalebox{0.81}{
\begin{tabular}{l c l c c c c c }
\hline  
\textbf{Model} & \textbf{Task} & \textbf{Setting} & \textbf{Test$1$ Acc} & \textbf{Test$1$ $F_1$} & \textbf{Test$2$ Acc} & \textbf{Test$2$ $F_1$} & \textbf{Avg Test Acc}\\ \hline
MatCha-chartqa-zero & C & zero & 29.1 & 46.4 & 29.3 & 48.3 & 29.2 \\
MatCha-plotqa1-zero & C & zero & 51.3 & 50.7 & 47.4 & 48.3 & 49.4\\
MatCha-plotqa2-zero & C & zero & 48.2 & 32.9 & 48.7 & 32.9 & 48.4\\
MatCha-finetune-class &	C & fine-tuning & 64.0 & 63.7 & 61.6 & 60.9 & 62.8\\
MatCha-finetune-multi &	M & fine-tuning & 59.4 & 59.1 & 61.1 & 60.9 & 60.2\\
MatCha-finetune-50/50-multi & M & fine-tuning & 61.9 & 61.2 & 63.0 & 62.1 & 62.4 \\
MatCha-chartqa-finetune-50/50-multi & M & fine-tuning & 59.1 & 58.8 & 61.4 & 60.7 & 60.2\\
MatCha-plotqa1-finetune-50/50-multi & M & fine-tuning & 60.6 & 60.5 & 61.6 & 61.4 & 61.1 \\
MatCha-plotqa2-finetune-50/50-multi & M & fine-tuning & 62.8 & 62.8 & 61.4 & 61.4 & 62.1 \\
\hline
DePlot-DeBERTa-class & C & fine-tuning & \textbf{75.0} & \textbf{75.0} & \textbf{72.5} & \textbf{72.5} & \textbf{73.8} \\
DePlot-TAPAS-class & C & fine-tuning & 61.6 & 61.0 & 60.4 & 59.7 & 60.8\\
DePlot-FlanT5-finetune-class & C & fine-tuning & 66.3 & 66.2 & 66.2 & 66.1 & 66.3 \\
DePlot-FlanT5-zero-multi & M & zero & 66.2 & 64.6 & 64.0 & 62.1 & 65.1 \\
DePlot-FlanT5-few-multi & M & few & 64.6 & 63.6 & 65.2 & 64.5 & 64.9\\
DePlot-FlanT5-finetune-multi & M & fine-tuning & 65.7 & 65.7 & 65.9 & 65.8 & 65.8 \\
\hline
GPT$3.5$ & M & zero & 59.0 & 56.2 & 51.0 & 48.7 & 55.0 \\
GPT$3.5$ & M & few & 56.6 & 52.8 & 61.0 & 59.5 & 58.8 \\
GPT$4$V & M & zero & 73.8 & 73.5 & 72.0 & 71.3 & 72.9 \\
GPT$4$V & M & few & 74.0 & 73.6 & 70.6 & 70.2 & 72.3 \\
\hline
Human baseline & C & - & & & & & \textbf{95.7} \\
\hline
\end{tabular}}
\caption{\label{tab:classification_results} Classification results for models trained on claim classification (C) and in a multi-task setting with explanation generation (M).}
\vspace{-2mm}
\end{table*}

\section{Experiments and Results}

We 
define the explainable fact-verification task against chart evidence as follows. 
Each dataset entry 
$(s_i, v_i, c_i, y_i, e_i)$ comprises a natural language statement $s_i$, a chart image $v_i$, a caption $c_i$ accompanying the chart image, a verdict label $y_i\in\{0, 1\}$, and an explanation $e_i$ in natural language. 
The statement $s_i$ is a claim that is either supported ($y_i = 1$) or refuted ($y_i = 0$) by the chart (Fig.~\ref{fig:chart_sample}).

\subsection{Baselines}

We evaluate several baselines 
on ChartCheck (Fig.~\ref{fig:baselines}), which can be grouped into two categories: 
chart-to-table (CTT)
and 
vision-language models (VLMs).
The CTT baselines extract charts' underlying table data before performing classification and explanation generation.
The VLMs are evaluated with Chain-of-Thought (CoT) prompting~\citep{Wei0SBIXCLZ22} in few- and zero-shot settings.


\paragraph{Experimental setup.}
Our baselines can be further grouped into single- and multi-task baselines. 
In the single-task setting, we first predict the claim verdict $\hat{y_i}$ and in a subsequent step generate the explanation using the claim, evidence, and either the predicted label $\hat{y_i}$ or gold label $y_i$ as input.
In the multi-task setting, we train/instruct models to jointly \textit{``classify and explain''}, resulting in output $(\hat{y_i}, \hat{e_i})$.
Inspired by the success of CoT on complex reasoning tasks, we also include the ChartCheck explanations in the input prompts as reasoning chains.\footnote{Prompts are given in the appendix.}
We evaluate models in three different training settings: finetuning, few-, and zero-shot.
The finetuned models are trained on the ChartCheck training set. 
For few-shot training, we provide selected training samples as input.

We evaluate claim classification using accuracy and macro $F_1$. We further evaluate explanations using 
BLEU~\citep{papineni-etal-2002-bleu}, ROUGE~\citep{lin-2004-rouge}, BERTScore~\citep{ZhangKWWA20}, METEOR~\citep{banerjee-lavie-2005-meteor}, and BLEURT~\citep{sellam-etal-2020-bleurt}. 

\begin{table*}
\centering
\scalebox{0.83}{
\begin{tabular}{l c l c c c c c}
\hline  
\textbf{Model} & \textbf{Task} & \textbf{Training setting} & \textbf{BLEU} & \textbf{Rouge} & \textbf{MET} & \textbf{BERTScore} & \textbf{BLEURT} \\ \hline
MatCha-finetune-multi &	M & fine-tuning & 17.1 & 37.3 & 38.3 & 67.8 & 0.48 \\
MatCha-finetune-50/50-multi & M & fine-tuning & 17.3 & 37.1 & 37.9 & 67.3 & 0.45 \\
DePlot-FlanT5-zero-multi & M & zero & 0 & 5.04 & 1.3 & 83.2 & 0.48 \\
DePlot-FlanT5-few-multi & M & few & 5.2 & 28.5 & 21.1 & 90.6 &  0.56 \\
DePlot-FlanT5-finetune-multi & M & fine-tuning & 17.3 & \textbf{46.3} & 36.0 & \textbf{91.5} &  0.50 \\
DePlot-FlanT5-explanation-gold & E & fine-tuning & \textbf{30.7} & 38.7 & 33.2 & 89.6 &  0.57 \\
DePlot-FlanT5-explanation-pre-labels & E & fine-tuning & 12.7 & 37.9 & 32.4 & 89.5 &  0.60 \\
GPT$3.5$ & M & zero & 10.0 & 32.3 & \textbf{42.9} & 89.1 & 0.43 \\
GPT$4$V & M & few & 10.3 & 33.6 & 41.7 & 89.4 & 0.39 \\
\hline
\end{tabular}}
\caption{\label{tab:explanation_results} Explanation generation results for testset $t1$ for model trained on explanation-only (E) and in a multi-task setting (M) with veracity classification.}
\vspace{-3mm}
\end{table*}

\noindent\textbf{$(1)$ Chart-to-table.}
The chart-to-table architecture solves the task in a three-step approach. Inspired by \citet{LiuEPKPLJCCA2022}, we decompose the chart-based fact-checking task into $(i)$ chart-to-table conversion, $(ii)$ table-based classification, and $(iii)$ explanation generation. 
First, for all charts we extract their underlying table data $T$ using the model DePlot~\citep{LiuEPKPLJCCA2022}. 
%
We evaluate four models on fact verification with $(c_i, v_i, T_i)$ as input: DeBERTa~\citep{HeLGC21},
TAPAS~\citep{DBLP:conf/acl/HerzigNMPE20},
FlanT$5$~\citep{Shen-etal-2023}, and GPT$3.5$.\footnote{For further details we refer to Appendix~\ref{sec:modeldetails_appendix}.}

\noindent\textbf{$(2)$ Vision-language models.}
For our vision-language baseline, we use \emph{MatCha}~\citep{liu2023} and GPT$4$-Vision (GPT$4$V)~\citep{OpenAI23}.
MatCha is a finetuned version of the Pix2Struct model~\citep{LeeJTHLEKWCT2022} for chart
reasoning. It achieves SOTA performance on multiple chart datasets, including ChartQA~\citep{DBLP:conf/acl/MasryLTJH22}, PlotQA v$2$~\citep{DBLP:conf/wacv/MethaniGKK20}, and the Pew charts subset of Chart-to-text~\citep{kantharaj-etal-2022-chart}. 
%
We render a language prompt directly on top of the chart image. 
We evaluate three MatCha models which are pretrained on different chartQA datasets.


\subsection{Results and Discussion}

\label{ssec:results_classification}

Table~\ref{tab:classification_results} and~\ref{tab:explanation_results} provide an overview of baselines' 
on claim classification and explanation generation. 

\paragraph{What baseline performs best on ChartCheck?}

\noindent \textbf{Claim Classification.}
\emph{DePlot-DeBERTa-class} achieves $75.0$ accuracy on $t1$ and $72.5$ on $t2$, outperforming other models.
The baseline translates chart images to tables, concatenates the table text into a text sequence, and uses the claim, table, and the chart caption as input to a DeBERTa large model 
fine-tuned on ChartCheck. 
While translating chart to table data is beneficial on our task, not all claims are verifiable based on the extracted tables, for example, if the claim refers to colors visible in the image.
Moreover, CTT models also propagate errors to subsequent stages resulting in factually-incorrect explanations if the predicted claim label was wrong. 
Among the VLMs, the best-performing model is GPT4V that achieves $73.8$.
However, both models lag far behind human performance of $95.7$.\\
\noindent \textbf{Explanation Generation.}
On explanation generation, no baseline outperforms across all metrics. Overall, MatCha models underperform on explanation generation compared to GPT-models and CTT baselines. 
\emph{DePlot-FlanT5-finetune-multi} outperforms other models on two out of five metrics. This model uses DePlot chart-to-table translation and finetunes FlanT5-base in a multitask setting on both claim classification and explanation generation. It performs best for testset $t1$ based on BERTScore ($91.5$) and ROUGE ($46.3$). For all other metrics, it scores among the top-$3$ baselines. We discuss the explanations' quality in more detail in Sec.~\ref{ssec:analysis}.

\begin{table}
\centering
\scalebox{0.75}{
\begin{tabular}{l c c c}
\hline  
\textbf{Chart Type} & \textbf{\#Claims} & \textbf{DePlot} & \textbf{MatCha} \\ \hline
Pie & 146 & 76.7 & 66.4 \\ 
Line & 122 & 62.2 & 59.0 \\ 
Bar & 104 & 81.7 & 64.4 \\ 
$3$D Pie & 39 & 66.7 & 39.0 \\ 
Area & 5 & 60.0 & 40.0 \\ 
Donut & 2 & 50.0 & 50.0 \\ 
\hline
\end{tabular}}
\caption{\label{tab:chart_attribute_performance} Performance of \textbf{DePlot}-DeBERTa-class and \textbf{MatCha}-finetune-class across different chart types.}
\vspace{-5mm}
\end{table}

\paragraph{What model insights do we gain through different training settings?}
%

\noindent \textbf{Pretraining.}
Pretraining MatCha on further chart datasets (e.g. chartQA), has 
no impact on the model's performance on ChartCheck. 
The redundancy of pretraining can be related to the datasets available for pretraining. 
Most of the questions in ChartQA/PlotQA only require retrieving a single value for the answer. ChartCheck claims' are complex and involve filtering, comparing, and commonsense knowledge among other reasoning types. 
Moreover, ChartCheck charts are more diverse (e.g. $3$D pie charts and area charts) compared to 
other datasets. 

\noindent \textbf{Multitask Learning.}
Training models jointly on classification and explanation generation shows no positive effect for claim classification.
While best-performing CTT and MatCha models are trained on classification-only, the multitask setting is more beneficial for explanation generation.
%

\noindent \textbf{Zero- and Few-shot Evaluation.}
Comparing \emph{DePlot-FlanT5-few-multi} to its zero-shot and finetuned equivalents, 
we find contrasting implications for claim classification and explanation generation. 
For claim classification, the accuracy for testset $t1$ and $t2$ either does not change or decreases (e.g. from $65.9$ to $64.0$).
However, prompting with few examples improves the model's explanation generation performance across all reported scores compared to its zero-shot counterpart. 
Finetuning with more examples has a larger impact on the generated explanations. Across all evaluation metrics, the few-shot model yields better explanations than the zero-shot model. Moreover, finetuning further improves the model's explanation
according to BERTScore and ROUGE. 
GPT models show similar claim classification and explanation generation performance for both few- and zero-shot settings.

\subsection{Analysis with Performance Breakdown}
\label{ssec:analysis}

\paragraph{Which charts pose a challenge for the models?}
Comparing model performance across \textbf{chart types} (Table~\ref{tab:chart_attribute_performance}), we observe that the CTT baseline performs best on bar charts while MatCha excels at pie charts. 
%
While the performance on line and donut charts is comparable, the performance difference increases for pie, bar, and area charts. This can be related to certain chart types easier translatable to table data.
%
CTT outperforms \emph{MatCha-finetune-class} on single-chart images reaching $73.0$ accuracy, whereas on \textbf{multi-chart images} the CTT baseline achieves only $50.0$ accuracy, less than MatCha ($58.3$). 
%
Interestingly, both baselines struggle with charts with  \textbf{legends}. The CTT baseline achieves $75.2$ on charts without legends and $65.3$ if a legend is given. Similarly, performance drops for the MatCha baseline from $65.2$ (no legends) to $56.8$ (legends). 

\begin{figure}
\centering
\includegraphics[scale=0.16]{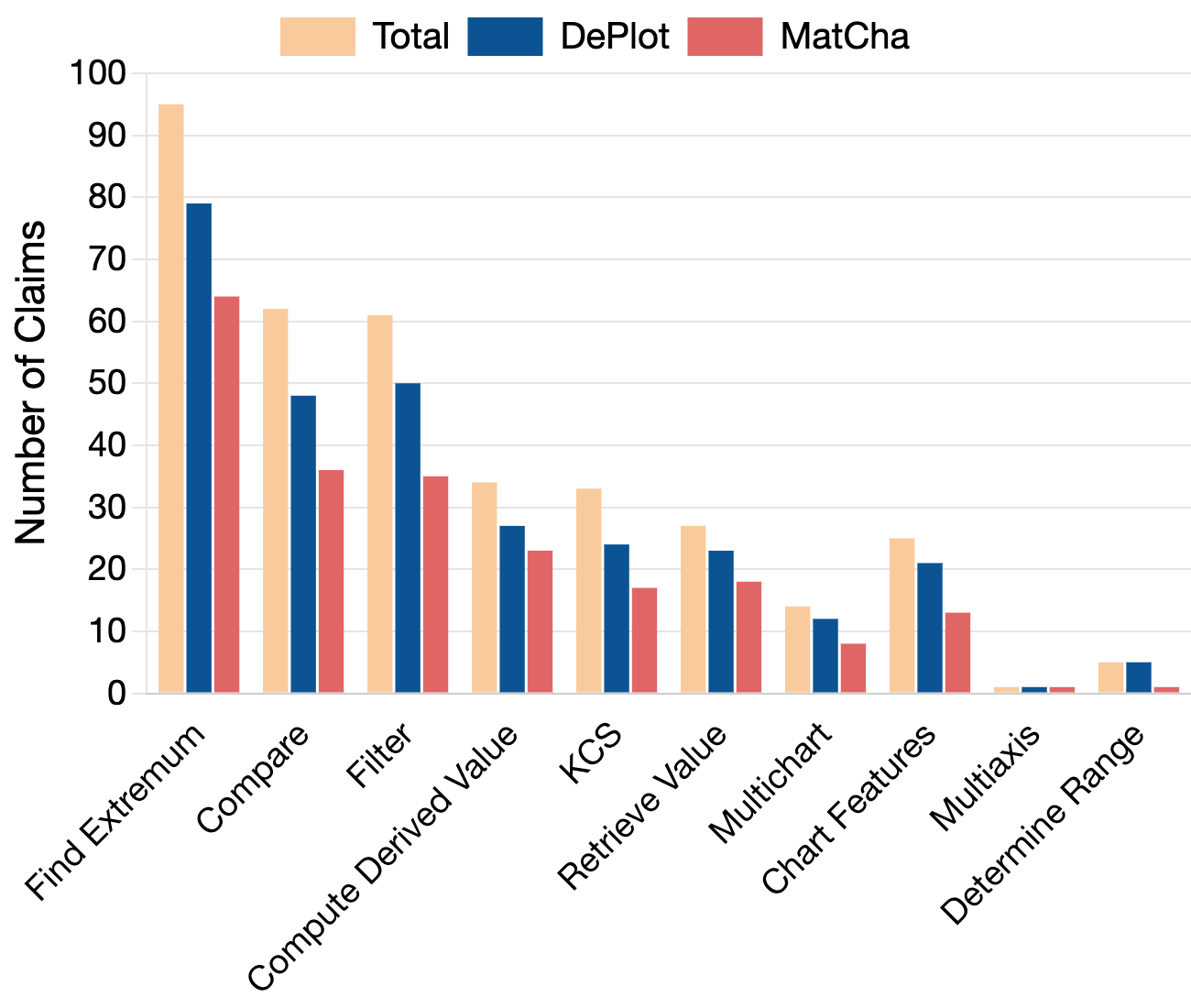}
\caption{\label{fig:reasoning_pred_count} Number of correct predictions by reasoning types in labelled subset with \textbf{DePlot}-DeBERTa-class and \textbf{MatCha}-finetune-multi.}
\vspace{-2mm}
\end{figure}

\paragraph{Do models struggle with specific types of chart reasoning?}
While both baselines variants struggle with certain reasoning types, in general the best-performing chart-to-table (CTT) baseline outperforms the best MatCha model across all reasoning types (see Fig.~\ref{fig:reasoning_pred_count}). 
The difference is most significant for the reasoning types \emph{determine range}, \emph{compare}, \emph{chart features}, and \emph{multichart}. CTT correctly predicts all claims requiring to determine a range of numbers correctly compared to the VLM reaching an accuracy of around $20.0$. For the other three reasoning types, CTT correctly predicts most claims (approx. $80.0$) 
while MatCha only achieves 
$60.0$ or less correctly. 
Both models struggle with claims requiring commonsense reasoning (\emph{KCS}).

\begin{figure}
\centering
\includegraphics[scale=0.65]{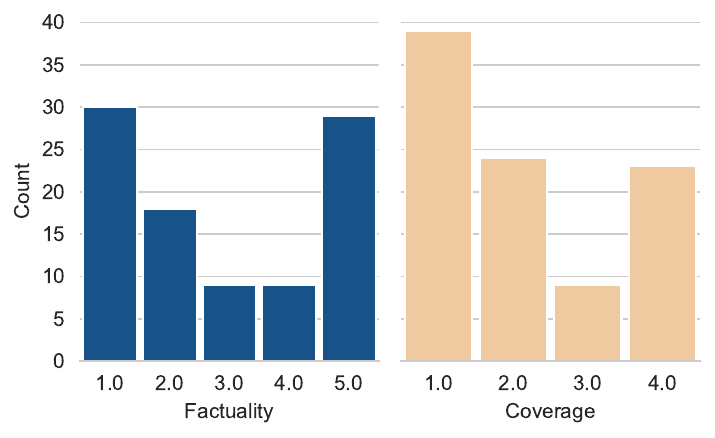}
\caption{\label{fig:explanation_eval1} Manual rating of model-generated explanations for factuality ($5$ implies totally correct) and coverage ($4$ implies reference explanation fully covered).}
\vspace{-3mm}
\end{figure}


\paragraph{What are common failure cases in chart-to-table translation?}
The charts in ChartCheck are extracted from the web and lack gold-tables required for automated evaluation. 
We manually evaluated a subset of DePlot-generated tables and found the following error cases:\footnote{Examples are given in the appendix.}
$(1)$~hallucinated labels for axes ticks and data points if they are not present or difficult to read;
$(2)$~errors in table translation of pie charts and grouped bar charts;
$(3)$~similar colors are confused if plotted against a non-white background;
$(4)$~missing or wrong titles;
$(5)$~incorrect text is extracted if background not white; 
$(6)$~generated tables sometimes repeat one row multiple times or miss some data points.

\paragraph{What are common failure cases in explanation generation?}

Rating model-generated explanations based on factual correctness and coverage of the reference explanation, we find that approximately half of the explanations are either completely incorrect (i.e. score $1$) or have major errors (score $2$; see Fig.~\ref{fig:explanation_eval1}).
A prevalent issue are explanations for refuted claims: models hallucinate explanations that contradict the content of the given chart but support the wrong claim. 
We also find explanations correctly mentioning the right claim label but giving a wrong explanation, ignoring the chart content. For example, a common issue is that models fail in reading numerical values from charts or in assigning them to the correct categories represented by a bar or a color in the image.
Moreover, explanations include reasoning errors such as stating that $45\%$ is \textit{``more than half''}. 
Expert annotators further rated explanations on a scale from $1-5$ for the categories \textit{Readability}, \textit{Coherence}, and \textit{Non-Redundancy}. More than $90\%$ of all explanations were rated \textit{``very well readable''}, \textit{``very coherent''}, and \textit{``no redundant text''}.\footnote{Further details in Appendix~\ref{sec:explanation_eval_appendix}}
Overall, these results imply that the model, while being able to generate very fluent, readable and coherent text, shows major limitations w.r.t. understanding and reasoning over charts to generate factually correct explanations.





\section{Related Works}

\paragraph{Datasets for evidence-based fact checking}
The goal of evidence-based fact checking is to verify a claim against evidence data. Typically, evidence is conveyed in different forms, e.g.  text~\citep{DBLP:conf/naacl/ThorneVCM18, augenstein-etal-2019-multifc, kotonya-toni-2020-explainable-automated}, tables~\citep{DBLP:conf/nips/AlyGST00CM21, akhtar-etal-2022-pubhealthtab, DBLP:conf/iclr/ChenWCZWLZW20}, and images~\citep{akhtar-etal-2023-multimodal}. 
%
While recent work introduced a first chart fact-checking dataset, \textit{ChartFC}~\citep{akhtar-etal-2023-reading}, the dataset has certain limitations. First, it is limited to bar charts, ignoring the diversity of data visualization found in real-world sources. 
Second, all ChartFC charts have been synthetically created using \textit{Python} visualization libraries. 
Moreover, ChartFC claims are extracted from the TabFact~\citep{DBLP:conf/iclr/ChenWCZWLZW20} dataset and therefore not written specifically for verification against charts.

\paragraph{Chart datasets for other tasks}
Chart question answering (ChartQA) is a closely related task to ours.
While existing ChartQA datasets have contributed to advancements in reasoning over data visualizations, most datasets have limitations in terms of size~\cite{DBLP:conf/icml/KimSK21}, template-based questions~\cite{DBLP:conf/cvpr/KaflePCK18}, synthetically generated charts~\cite{singh-shekhar-2020-stl} or missing explanations. 
While ChartQA~\citep{masry-etal-2022-chartqa} addresses many of these limitations, some challenges remain.
%
First, ChartQA images are collected from four 
websites 
and while they represent real-world scenarios, the charts are created by professionals, 
thus omitting 
the noise, design best-practise violations, and other issues found in charts by non-professional practitioners. 
Our dataset includes annotations for reasoning types and detailed visual attributes (e.g. presence of datapoint labels, 
legends, axis titles or
background grid characteristics), 
which are absent in comparable datasets.
%
%
Finally, ChartCheck is the first chart dataset with explanations, which has several benefits, such as better model training (i.e. allowing models to understand the reasoning behind a classification label), robustness, 
and transparency. 
This can further aid in building explainable models.

\paragraph{Modeling approaches for charts}
Various approaches have been developed for chart modeling. 
One direction encodes both chart image and text, and applies a fusion method (e.g. attention) to combine the features into a joint representation~\citep{
DBLP:conf/acl/MasryLTJH22, DBLP:conf/wacv/KafleSPCK20}.
Another popular approach is to first extract the chart's table data, before applying table models~\cite{DBLP:conf/chi/KimHA20, LiuEPKPLJCCA2022, DBLP:conf/wacv/MethaniGKK20}.
Finally, more recent models use visual encoders to encode charts and their context and language decoders for text generation and solving NLP tasks, e.g. question answering~\citep{masry-etal-2023-unichart, ye-etal-2023-ureader, LiuWYCSCYY23, XiaZPLYSYQ23}. We evaluate ChartCheck on \textit{DePlot}~\cite{LiuEPKPLJCCA2022}, \textit{MatCha}~\citep{liu2023}, and \textit{GPT4-Vision}~\citep{OpenAI23} as SOTA models of each category.



\section{Conclusion}

We introduce ChartCheck, the first dataset for explainable fact-checking with $1.7k$ real-world chart and $10.5k$ human written claims and explanations. 
We evaluate SOTA models, including chart-to-table baselines and VLMs, in a few-/zero-shot and finetuned setting. Our best baseline yields 
$73.8$ accuracy, lagging far behind human performance. 
Moreover, we study reasoning types and visual attributes that pose a challenge to SOTA models, as well as failures related to model-generated explanations.


\section*{Limitations}

The presented work exhibits the following limitations. First, the daatset is limited to English-only, whereas fact-checking misinformation is a global need that requires solutions in multiple languages. Further research is needed to extend chart-checking beyond English.
Second, although ChartCheck includes a wide range of charts, certain chart types such as heat maps, scatter plots, and histograms are missing. Addressing these chart types should be a focus of future research.
Finally, in our attempt to include diverse charts from real-world sources, we were unable to obtain gold table data for the charts. In the future, it would be beneficial to construct a dataset that includes diverse charts along with their underlying table data. This would enable the study of a broader range of models and their performance on such data. It would be also interesting to make the data three way labeled including "uncertain" label in addition to "true" and "false" claim.


\section*{Ethics Statement}

This paper introduces a dataset for automated fact-checking against chart images. The dataset is intended for research purposes, not as the sole means of evaluating real-world applications. The labels (i.e. supports and refutes) describe a claim's
veracity given the evidence table. We do not make
any statement on ChartCheck claims' truthfulness in a real-world context.

Prior to the crowdsourcing, we obtained ethical clearance from the relevant authority in our institution. We informed the annotators about all data being collected and its purpose. Participants had the opportunity to withdraw at any
time and to provide feedback at the end of each task.
All annotators were from English speaking countries.
The payment was above the minimum wage and
decided based on the time workers spent on the
pilot tasks.
In order to support future research, we plan to release all the scripts and resources used for dataset creation and model evaluation. This will facilitate and encourage further research in this field.

\section*{Acknowledgements}

We thank Fangyu Liu for very helpful coversations and feedback in earlier stages of
this project.

\bibliography{anthology,custom}

\appendix

\appendix

\section{Future Challenges}






Based on our experiments and results, we observed the following challenges as promising directions for future research:
First, there is a need for models that can both extract (visual) information from charts and successfully solve chart-related tasks that involve various reasoning skills. Although our DePlot-based baselines performed well, important visual information is lost during image-to-table translation.
Second, there is a need for future datasets that expand on this work across various dimensions. One direction to explore is the creation of multilingual chart datasets. Another direction can consider fact-checking against data visualization types that are not currently included in ChartCheck, such as maps and infographics.
Studying charts within the broader context of documents is another interesting direction to pursue. In real-world settings, charts are often embedded within documents. Evaluating claims about charts may require combining information from various sources, such as text, tables, or other images found alongside the chart.

\section{Chart Misinformation}

\begin{figure}
\centering
\includegraphics[scale=0.62]{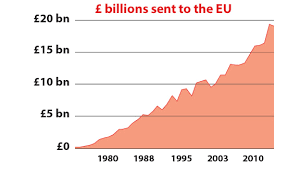}
\caption{\label{fig:chart_misinfo_sample2} Chart distributed during the Brexit campaign.}
\end{figure}

\begin{figure}
\centering
\includegraphics[scale=0.24]{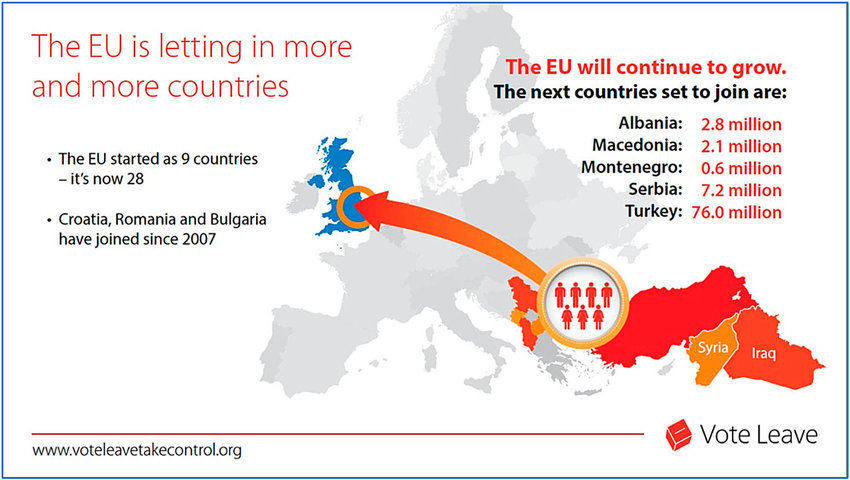}
\caption{\label{fig:chart_misinfo_sample3} Chart distributed during the Brexit campaign.}
\end{figure}

Fig.~\ref{fig:chart_misinfo_sample2} and~\ref{fig:chart_misinfo_sample3} are examples of chart-backed arguments spread during the Brexit referendum.\footnote{\url{http://www.voteleavetakecontrol.org/why_vote_leave.html}}

\section{Dataset Collection}
\label{sec:appendix_dataset}

\subsection{Wikimedia Commons Repository}
\label{ssec:dataset_details}

\begin{figure}
\centering
\includegraphics[scale=0.38]{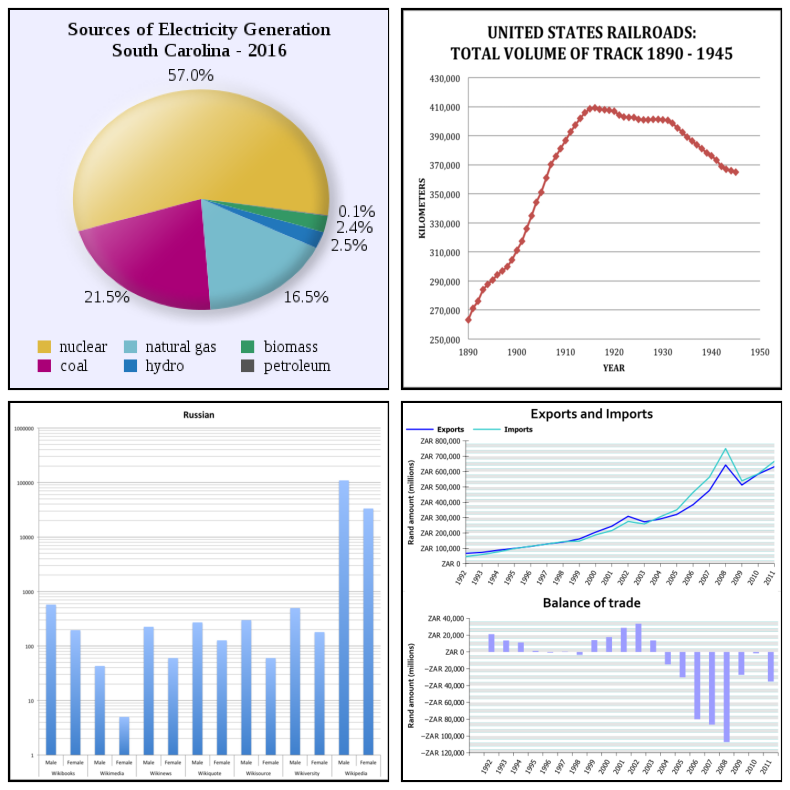}
\caption{\label{fig:chartcheck_charts} Examples of charts in the ChartCheck dataset.}
\end{figure}

\begin{figure*}
\centering
\includegraphics[scale=0.35]{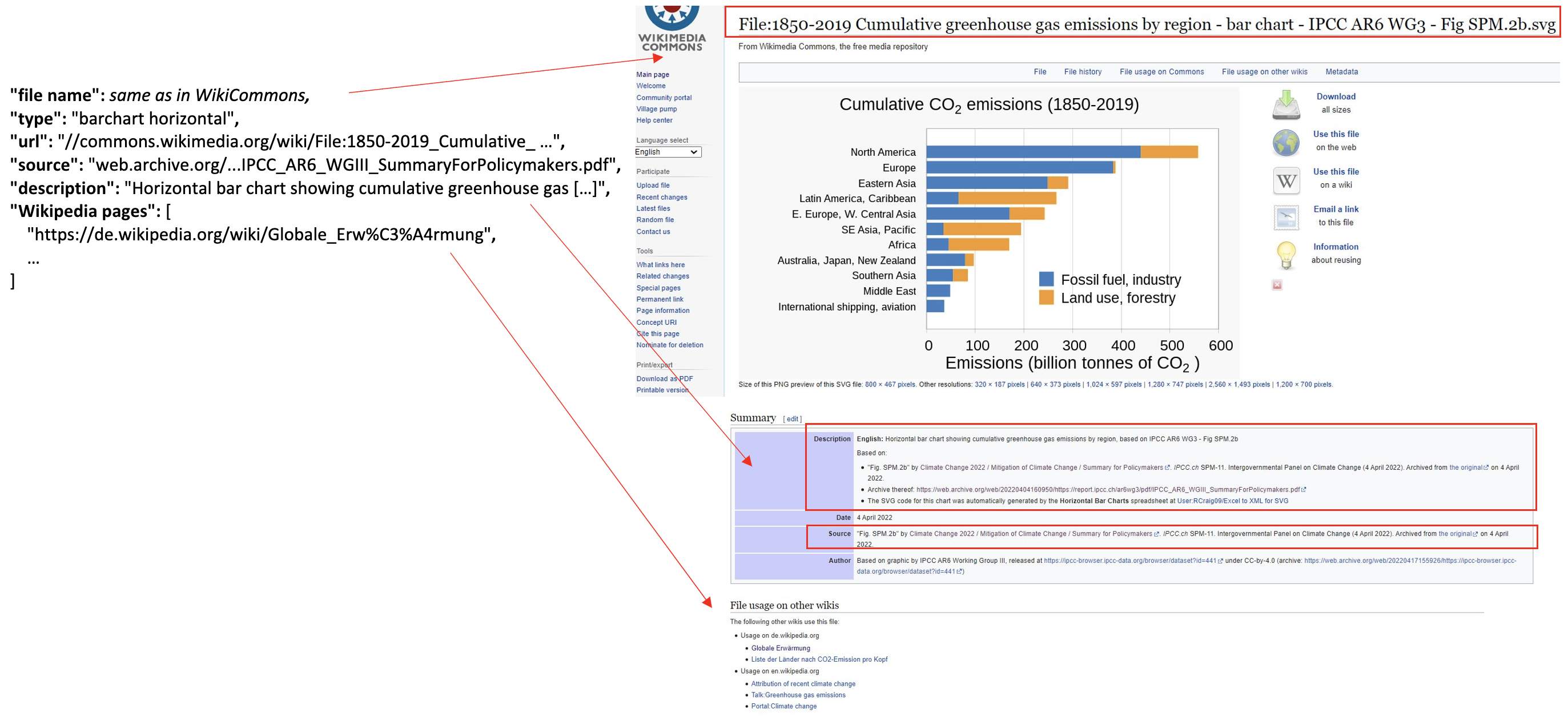}
\caption{\label{fig:wikimedia} Chart example from the Wikimedia Commons page.}
\end{figure*}

We collected charts using Wikimedia Commons, a project of the \emph{Wikimedia Foundation} (similar to  Wikipedia and WikiData).\footnote{\url{https://commons.wikimedia.org/wiki/Commons:Welcome}} 
Wikimedia is a media repository of images, videos, and audios which are used across all Wikimedia projects and not subject to copyright restrictions. As of May 2023, the repository contained over $93$ million files.
In Wikimedia Commons, images are labelled with categories which are part of a multi-hierarchy structure of categories.
We used Wikimedia's API and the categories \emph{``Horizontal\_bar\_charts''}, \emph{``Vertical\_bar\_charts''}, \emph{``Line\_charts''}, \emph{``Pie\_charts\_in\_English''} (e.g. chart in Fig.~\ref{fig:chart_sample}), and \emph{``Scatterplots''} to extract chart images from the repository. See Fig.~\ref{fig:wikimedia} for a Wikimedia Commons entry. 
We only extracted charts, if their description was in English. We used the spaCy~\citep{honnibal2020spacy} language detection tool.
Additionally, we extracted the file name, category, description, url, source (if available), and the Wikipedia pages the image is linked to. 

\subsection{Crowdsourcing}
\label{ssec:crouwdsourcing_appendix}

\paragraph{Mechanical Turk Details.}
For the first and third crowdsourcing task, we created sets of seven annotation tasks each, out of which two were from a gold-labelled sample set annotated by the authors. 

For the second annotation task (i.e. claim and explanation generation), we provided annotators a set of seven charts and captions per taskset. For each chart-caption pair, we asked them to write one claim which was supported by the chart and its caption and one which was refuted, as well as an explanation.
The claim length was restricted to $[5; 30]$ tokens and the explanation text to $[10; 100]$ tokens. To improve the submission quality, we implemented automated checks on the claim and explanation text. 

\paragraph{Quality Assurance.}
We included gold annotated samples in the first and final crowdsourcing task. We created a gold standard of $50$ charts for the chart filtering task and $40$ for the verification task. We used two gold samples per task. 
Workers who failed those two samples could not submit their work. 
For the second annotation task (i.e. claims and explanation generation), we manually evaluated one claim and explanation per submitted task set, before accepting the work. We banned workers with malicious behaviour, e.g. workers who submitted the same claims as supporting and refuting example.
Automatic check evaluated claims and explanations for their token length, use of ambiguous wording (e.g. mostly, perhaps, and some), and negation words in case of refuting claims/explanations.  



\paragraph{Training and recruitment of workers.}
We allocated the chart filtering and validation task to three crowdworkers each. Workers with a record of min. $1000$ previously-approved tasks and an approval rate of min. $95\%$ were eligible. All workers first passed a chart literacy qualification test. 
We included examples of
expert-labelled tasks in the instructions and rationales for the chosen labels.

\begin{figure}
\centering
\includegraphics[scale=0.24]{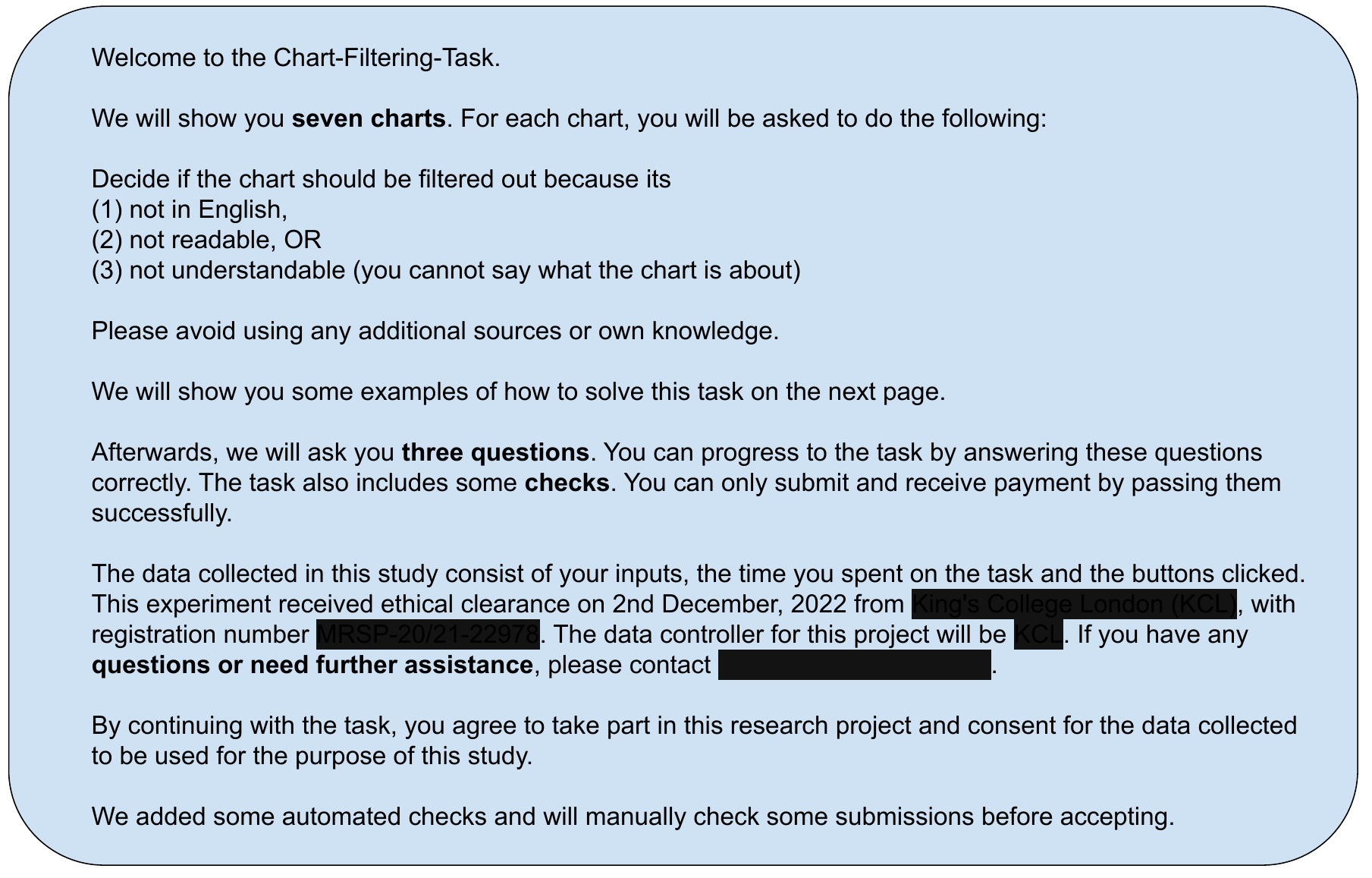}
\caption{\label{fig:chart_filtering_instruction} Instructions we showed annotators at the start of the chart filtering task.}
\end{figure}

\begin{figure}
\centering
\includegraphics[scale=0.25]{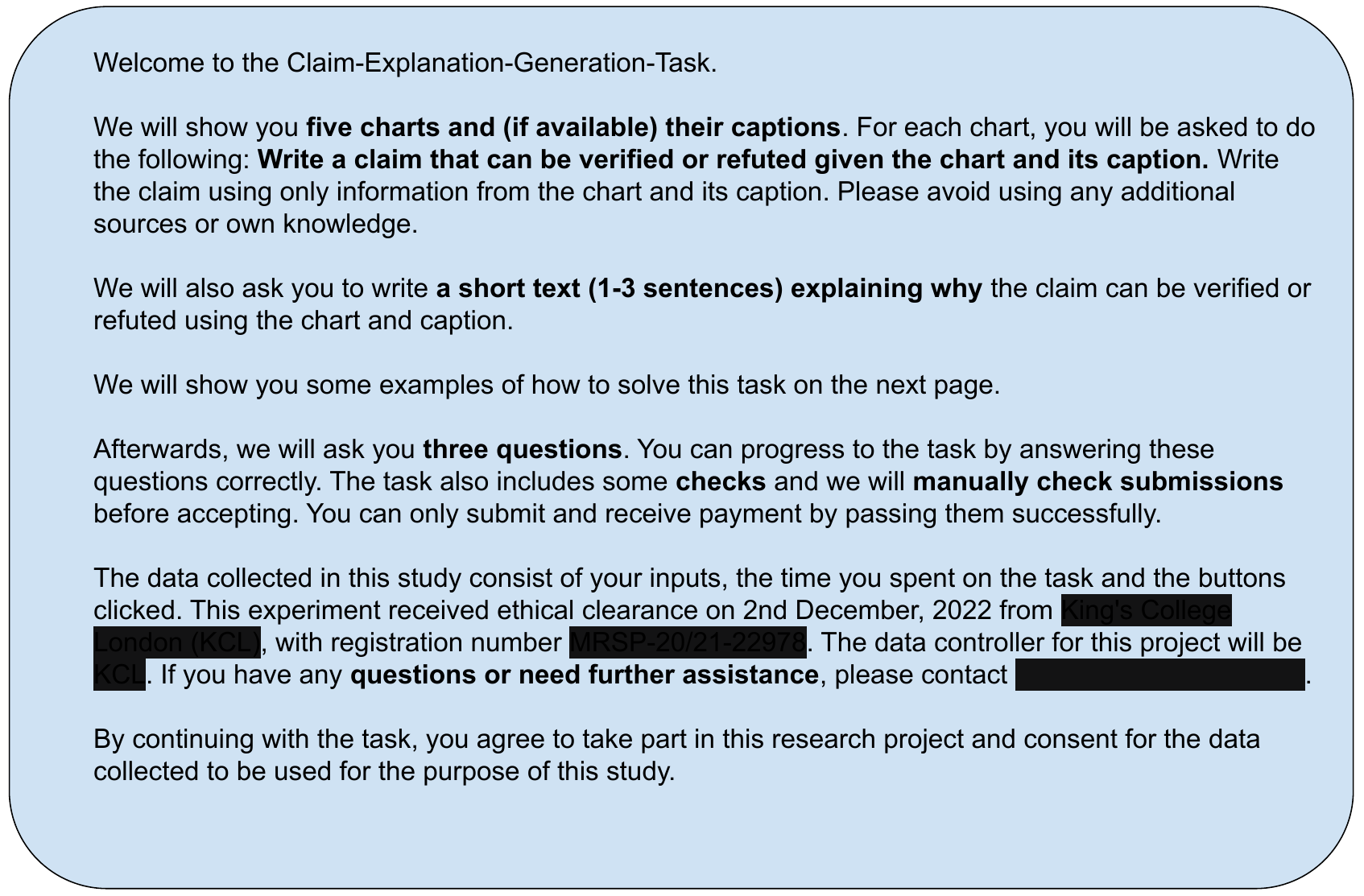}
\caption{\label{fig:claim_generation_instruction} Instructions we showed annotators at the start of the claim/explanation generation task.}
\end{figure}

\begin{figure}
\centering
\includegraphics[scale=0.25]{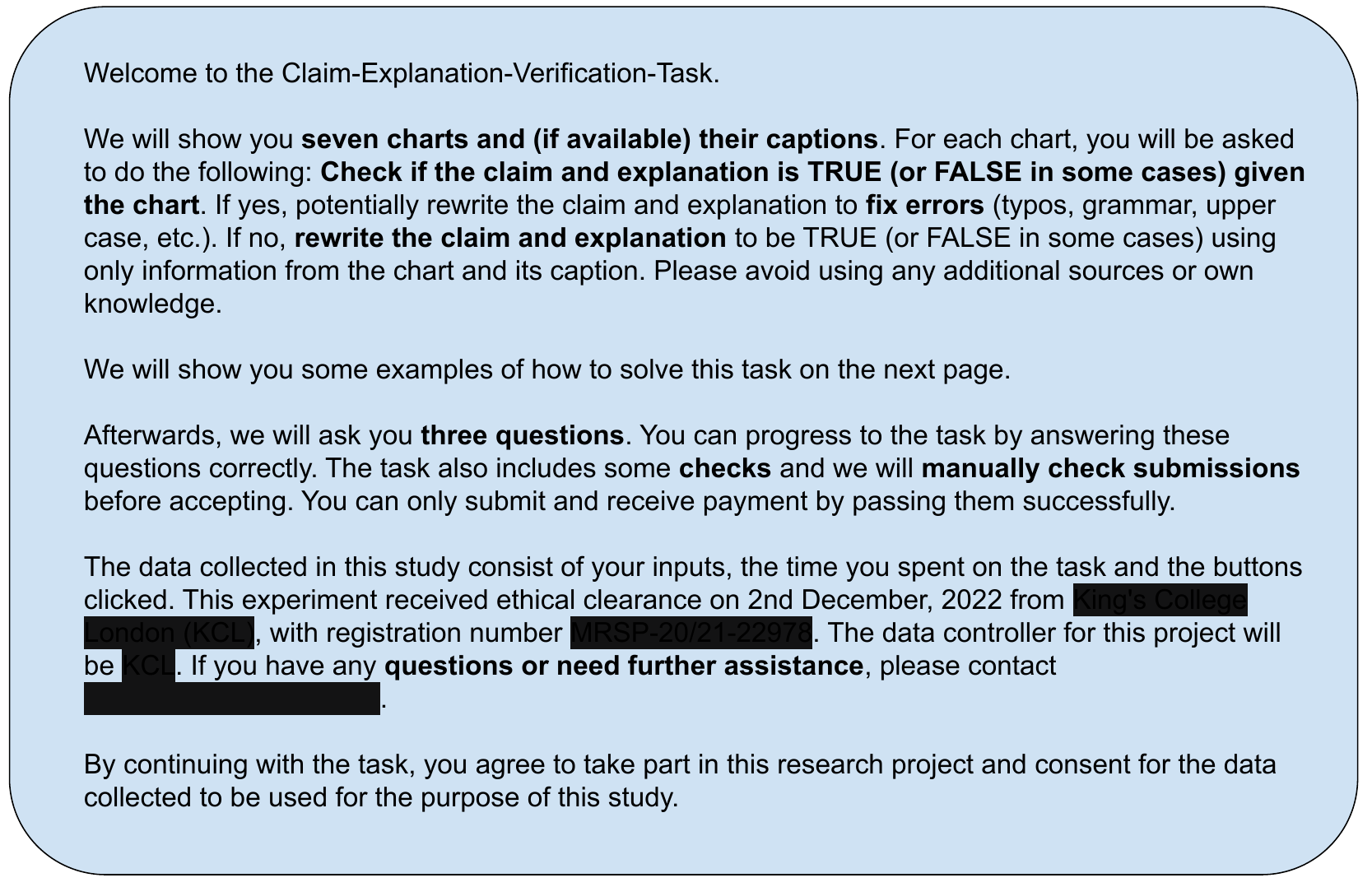}
\caption{\label{fig:claim_verification_instruction} Instructions we showed annotators at the start of the claim/explanation verification task.}
\end{figure}

Figure~\ref{fig:chart_filtering_instruction}, \ref{fig:claim_generation_instruction}, and \ref{fig:claim_verification_instruction} depict instructions presented to annotators at the beginning of the chart filtering, claim/explanation generation, and verification tasks, respectively. 

\subsection{Dataset Examples}

Fig.~\ref{fig:chartcheck_charts} shows examples of charts in ChartCheck.

\section{Modeling Details}
\label{sec:modeldetails_appendix}

\paragraph{Technical details.}
Before using images as input to models we converted a subset of images which were RGBA images to RGB images by adding a white chart background. 
We trained all MatCha models with AdaFactor optimizer, a learning rate of $1e-5$, a batch size of $4$ and a max patch size of $2048$ for $15,000$ step. 
A checkpoint was saved every 300 step and the model with the best validation loss was selected. 
We used a max patch size of $2048$.
For DeBERTa-based classification, we evaluated and compared two models, the DeBERTa base model and a model pretrained on various NLI datasets (e.g. MNLI~\citep{williams-etal-2018-broad} and FEVER~\citep{DBLP:conf/naacl/ThorneVCM18}).
We selected the NLI-pretrained DeBERTa model due to better performance and fine-tuned it with a learning rate of $1e-06$ and a batch size of $4$. 
We fine-tuned a TAPAS model which was previously trained on the table fact-checking task TabFact~\citep{DBLP:conf/iclr/ChenWCZWLZW20}.
The FlanT$5$ base model was fine-tuned on ChartCheck with a learning rate of $5e-5$ and batch size of $8$.

\paragraph{Instructions for Model Prompting.}
For VLM experiments, our prompts consist of an instruction to the model with the claim (\textit{``[claim]''}) and caption (\textit{``[caption]''}) rendered on top of the input image. For claim classification tasks, we used the following prompt: \textit{``The chart below has the following caption: [caption]. Given the caption and the chart below, is the following claim true: [claim]?''}. For claim classification with explanation generation task, we used the following prompt: \textit{``The chart below has the following caption: [caption]. Given the caption and the chart below, is the following claim true: [claim]? Explain why?''}. 

\paragraph{Chart-to-table baseline details.}
We finetune the first three models on the ChartCheck training data and evaluate GPT$3.5$ in a few- and zero-shot setting using CoT prompting.
Finally, we finetune the FlanT$5$~\citep{Shen-etal-2023} base model 
for explanation generation as described above.
We also test FlanT$5$ in a multi-task setting (i.e. claim classification and explanation generation). We evaluate a finetuned, (\emph{DePlot-FlanT5-finetune-multi}), few-shot, (\emph{DePlot-FlanT5-few-multi}), and zero-shot (\emph{DePlot-FlanT5-zero-multi}) version of the multitask FlanT$5$ model.

\begin{figure*}
\centering
\includegraphics[scale=0.63]{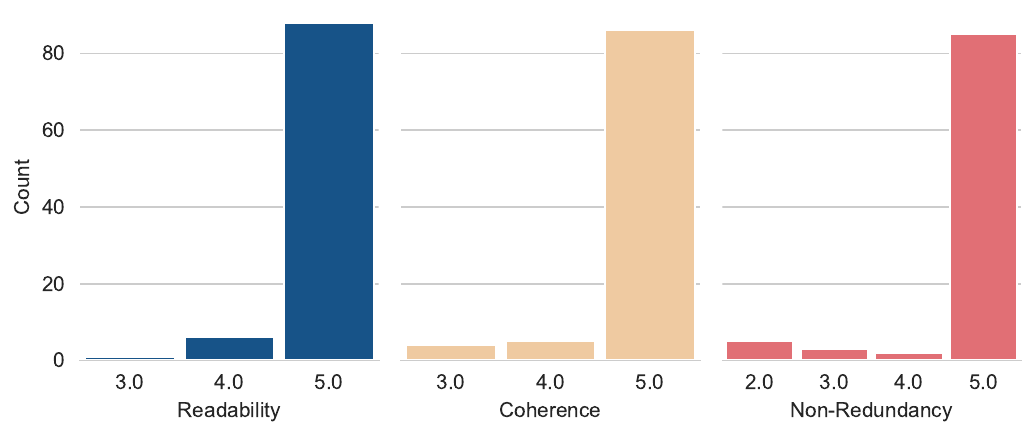}
\caption{\label{fig:explanation_eval2} Results of manual explanation evaluation for the categories \textit{Readability}, \textit{Coherence}, \textit{Non-Redundancy}. Expert annotators rated model-generated explanations on a scale from $1-5$ for these categories. Score $5$ depicts \textit{``very well readable''}, \textit{``very coherent''}, and \textit{``no redundant text''} for the given categories.}
\end{figure*}

\paragraph{Vision-langauge model baseline details.}
We evaluate three MatCha models in a zero shot setting on our testset: MatCha fine-tuned on ChartQA (\emph{MatCha-chartqa-zero}), PlotQA v$1$ (\emph{MatCha-plotqa1-zero}), and PlotQA v$2$ (\emph{MatCha-plotqa2-zero}). 
In a fine-tuned setting, we have MatCha fine-tuning only on the classification task (\emph{MatCha-finetune-classification}) and in a multi-task setting (\emph{MatCha-finetune-multi}). 
Moreover, we experiment with a $50/50$ training data split during fine-tuning (\emph{MatCha-finetune-50/50-multi}). For the first half of the dataset, we only fine-tune with classification prompts and for the second half we include explanations as well to improve overall classification performance.

\section{Explanation Evaluation}
\label{sec:explanation_eval_appendix}

Fig.~\ref{fig:explanation_eval2} shows the results of manual explanation evaluation for the categories \textit{Readability}, \textit{Coherence}, \textit{Non-Redundancy}. Expert annotators rated $100$ model-generated explanations on a scale from $1-5$ for these categories. Score $5$ depicts \textit{``very well readable''}, \textit{``very coherent''}, and \textit{``no redundant text''} for the given categories.

\section{DePlot: Chart-to-table examples}
\label{sec:appendix_deplot} 

\begin{figure}
    \begin{tabular}{ c c }
        \fbox{\begin{minipage}{18.6em}
                \small
                \centering
                \raggedright{\textbf{Chart:}} \\
                \centering
                \includegraphics[scale=0.3]{figures/chart_to_table.png}\\
                \rule{\linewidth}{0.01em}
                \raggedright{\textbf{Table:}}\\ | TITLE | Frequency of mobile Wikipedia usage\\
Frequency of mobile Wikipedia usage | in | us\\
At least once a day | 29 | 13\\
At least once a month | 9 | 15\\
At least once a week | 36 | 45\\
Less than once a month | 6 | 10\\
I never read Wikipedia articles on a mobile device | 5 | 7
            \end{minipage}} &
    \end{tabular}
    \caption{\label{fig:charts_translation1} Example of DePlot Chart-to-Table conversion}
\end{figure}

\begin{figure}
    \begin{tabular}{ c c }
        \fbox{\begin{minipage}{18.6em}
                \small
                \centering
                \raggedright{\textbf{Chart:}} \\
                \centering
                \includegraphics[scale=0.4]{figures/chart_to_table2.png}\\
                \rule{\linewidth}{0.01em}
                \raggedright{\textbf{Table:}}\\ TITLE | Sources of Electricity Generation\\
New Mexico - 2018\\
Source of Electricity Generation\\
New Mexico - 2018 | coal | wind | other\\
coal | natural gas | solar \\
 2018 | 0.407 | 0.351 | 0.351 | 0.351 | 0.007 \\
 2019 | 0.407 | 0.351 | 0.351 | 0.351 | 0.007\\
            \end{minipage}} &
    \end{tabular}
    \caption{\label{fig:charts_translation2} Example of DePlot Chart-to-Table conversion}
\end{figure}

\begin{figure}
    \begin{tabular}{ c c }
        \fbox{\begin{minipage}{18.6em}
                \small
                \centering
                \raggedright{\textbf{Chart:}} \\
                \centering
                \includegraphics[scale=0.25]{figures/chart_to_table3.png}\\
                \rule{\linewidth}{0.01em}
                \raggedright{\textbf{Table:}}\\ TITLE | United States Female Arson Arrests\\
Age | Number arrested\\
9 | 0.36\\
10 | 0.82\\
13 | 2.74\\
15 | 3.11\\
16 | 1.83\\
17 | 2.01\\
18 | 1.27\\
19 | 1.44\\
20 | 1.31\\
21 | 1.52\\
22 | 1.62\\
23 | 1.80\\
24 | 1.67\\
25 | 1.69\\
30 | 1.39\\
35 | 1.35\\
40 | 1.33\\
45 | 0.99\\
50 | 0.72\\
55 | 0.52\\
60 | 0.23\\
65 | 0.11\\
            \end{minipage}} &
    \end{tabular}
    \caption{\label{fig:charts_translation3} Example of DePlot Chart-to-Table conversion}
\end{figure}

Figures~\ref{fig:charts_translation1},~\ref{fig:charts_translation2}, and~\ref{fig:charts_translation3} show ChartCheck images and extracted table data.

\end{document}